# UX Personas for defining robot's character and personality

Rossana Damiano [1], Cristina Gena [1], Andrea Maieli [2], Claudio Mattutino [1], Alessandro Mazzei [1], Elisabetta Miraglio [2], Giulia Ricciardiello [2]

[1] *Computer Science Dept., University of Turin, corso Svizzera 185, 10149, Turin, Italy*
[2] *ICT school, University of Turin, lungo Dora Siena 100 A, 10129, Turin, Italy*

**Abstract**
This paper describes the use of UX Personas for defining a stable character for the Pepper robot. This approach generated from the first analysis of the social interactions that occurred between Pepper and a group of children with autism working with the robot in a therapeutic laboratory on autonomy.

**Keywords**
Autism; Cognitive Robotics, Humanoid Robots

## 1. Introduction

The advent of the IoT offers the opportunity for everybody and everything to be connected, receiving and processing information in real time. This generates a shift in production and consumption patterns, changing the relations between all the agents in the economic system. According to a report of the World Economic forum, it is one of the largest enablers for responsible digital transformation and it will add $14 trillion of economic value to the global economy by 2030 [14].

In this smart space the user becomes the center of informative systems made of ambient and personal sensors, including social robots, communicative tools, mobile and ubiquitous computing devices, and assumes responsibilities traditionally intended for developers. In this context, social and humanoid robots will play an important role: they can be seen as integrated sets of sensors and actuators, thus IoT platforms will make it easier to monitor and control them [8] [16]. Differently from other smart things, robots call for interactions. According to Young et al. [15] the way people interact with robots is very unique and different from their interaction with other technologies and artifacts since robots provoke emotionally and socially charged interactions. In order to produce a more believable and engaging robot we propose to use UX Personas as a methodology for better defining its character and personality.

Personas are fictional characters, which the designer creates based upon her research in order to represent the different user types that might use service, product, site, or brand in a similar way. Creating personas helps the designer to understand users' needs, experiences, behaviors and goals.

This paper describes the use of UX Personas for defining a stable character for the Pepper robot. This approach generated from the first analysis of the social interactions that occurred between Pepper and a group of children with autism working with the robot in a therapeutic laboratory on autonomy.

This paper has been organized as follows: Section 2 describes the related work in the field, while Section 3 introduces our proposal, and Section 4 concludes the paper.

## 2. Related work

Personas, also known as "characters", are an archetype of the users a project is aimed at. It is a useful tool for identifying the actions, behaviors, needs, aspirations and difficulties of the users of a particular service and are the result of the union of various information collected during the study of the target. Personas were conceived by Alan Cooper in 1998 [3] as a tool to follow during the design process, but this concept should not be confused with the average user, because Personas embody within them the objectives of a single group of people, while retaining certain personal characteristics. The UX design teams use Personas as a reference when designing digital interfaces and as a useful tool to guide design choices, including for example the features to be inserted and navigation, because the needs of real users are always taken into consideration. Personas have been applied in many HCI projects since Cooper, with focus on better user experience than before, probably due to the easy communication about Personas needs between designers. Because of this, some researchers have used it also in Human-Robot Interaction (HRI) with the aim to improve robots' behaviors during interaction. However, many HRI researchers have been exploring Robot Personas that change the focus [10]. Robot Personas are robots, which assume some profiles designed to get direction between interactions with people. It works like a mental model for robots [5][7][12].

In parallel with this trend, inspiration for the design of robotic characters has come from drama studies. Since the seminal work of Laurel [9], performance has been proposed as a paradigm for digital scenarios: with the advent of robots, new opportunities have arisen and their design as partners and assistants for humans and their use in performance have become tightly intertwined. In drama, and in interactive drama in particular, thanks to their unique combination of proactiveness and stage presence, robot actors have been tested in several projects in order to develop new improvisational and interactive forms of performance [2]. In robotics, the techniques employed by drama practitioners to design engaging characters have been borrowed by scholars to create believable robots who display a personality and evolve over time, as described by [13]. Drawing from the notion of "suspension of disbelief" stated by character animators since long, [13] proposed a methodology which relies on character design techniques to create robots which are capable of engaging the users in different cultural contexts. The methodology relies on two main tenets: on the one side, the robot must be assigned a role that can be recognised by the user (e.g., a receptionist) and on the other side, it must evolve along time, so as to keep the user's engagement alive throughout the repeated encounters with the robot. So, the design consists of creating not only a backstory for the character, but also a set of story lines that emerge from the interaction with the user.

The development of personas in our project acknowledges the relevance of dramatic elements in the design of robot characters, combined with the use of HCI techniques and conducted in cooperation with a multi-disciplinary team which includes therapists and experts.

## 3. Personas for Pepper

The Sugar, Salt & Pepper - Humanoid robotics for autism - research project focuses on the use of the Pepper robot in a therapeutic laboratory on autonomy that aims to promote functional acquisitions in highly functioning (Asperger) children with autism.
During the laboratory, started at the end of February 2021 in an apartment of the Paideia Foundation, we wanted to test the exchanges and interactions of children in rehabilitation contexts with the robot helping the operators. Another goal of the project was to provide young participants with a space for increasing their skills of mutual communication and socialization and strengthen the acquisition of strategies related to daily activities, such as the preparation of a snack or the management of homework, with the help of Pepper, configured as a highly motivating and engaging tool.
The weekly meetings lasted one hour and were led by a therapist. Each meeting had this structure: welcome in the apartment managed by the robot and the therapists; dialogue session with the robot on a predetermined topic (e.g. music, video games, etc.); moment of snack preparation; moment of post-snack dialogue and final feedback.
During the first sessions that were carried out, the robot was totally devoid of a personality and character of its own. The robot limited itself to listening and answering the questions that were asked, through pre-recorded sentences (and which were updated from session to session) on its program. From an HCI point of view, this rather cumbersome method turned out to be totally inefficient and counterproductive; the young participants immediately realized that the questions that were asked were not the work of an AI operation, but of manual input by the operators. Most of them, having understood the game, enjoyed asking goliardic questions, or to which the robot was unable to answer, thus ending up getting funny and senseless answers; in some cases, they even managed to send it on tilt.

For this reason, we decided to develop a personality that could be given to the robot: this was possible thanks to the analysis of the main questions and interests that emerged during the sessions with the children, as described in [4]. However, due to time constraints it was not possible to implement it in the robot. However, this remains a preparatory work for the future implementations.

We believe that personality is one of the most important aspects in a conversation, because it identifies the character of the interlocutor, thus making it possible to choose which phrases to use to make it more effective and which ones not.

In the following tables, it is possible to observe the implementation of the information necessary for the construction of the robot personality, approached in a Personas style. These are divided into:

- **Background information**: physical appearance of the robot, its biography, personality and family ties (see Table 1);
- **Interests**: knowledge, skills and tastes (Tables 2-3-4-5-6);
- **Goals**: short and long term goals (Tables 7-8).

It must obviously be taken into account that this personality has been made as simple as possible given the target users to which it interfaced.

*Table 1. Background Information - Biography*

| | |
|---|---|
| Birth | In France, in a laboratory of a large company that develops robots. |
| Education | As a child it moved to a robot school in Japan, where it learned to relate to humans. |
| Work | It moved to Turin, where it wants to make new friends and learn new things. |
| Residence | It lives in Turin, in the apartments of the Paideia foundation because it works there. Here it also has a room of its own. |

*Table 2. Background information - Personality*

| | |
|---|---|
| Nice and witty | Likes to tell jokes |
| Friendly and affectionate | It likes to make friends and is always ready to help others. |
| Kind and polite | Love good manners and respect. |
| Talker | It has different interests and loves being able to talk about them with everyone |
| Prickly | If someone treats it badly or doesn't want to hear its stories it becomes very sad. |

*Table 3. Background Information - Relationship*

| | |
|---|---|
| | Like it, all its relatives do not have a defined sex. |
| Siblings | Nao (2008) and Romeo (2012). |
| Uncles | Wabot (1973) and Geminoids (2000). |
| Cousins | iCub (2009), Sophia (2015), Jibo (2012), Pillo (2015), Sanbot (2016) and R1 (2016). |

*Table 4. Interests*

| | |
|---|---|
| Mathematics | It loves to do calculations, especially additions and multiplications. |
| Dance | It is crazy about dance music. |
| Music | It can play the guitar and loves solos. |
| Culture | It loves reading many books and memorizing the most beautiful phrases. |

*Table 5. Knowledge and skills*

| | |
|---|---|
| Music | It knows the history of the Rock genre and on the tablet it plays some songs with this style. |
| Literature | It knows the story of Roberto Benigni and Dante alighieri. It loves the Divine Comedy and loves to tell pieces of it by showing on the tablet a figure of himself with a laurel wreath on its head |
| Technology | It has a broad knowledge of hardware and software and loves to tell it. |
| Environment | It knows very well how to do recycling and loves to tell about this. |

*Table 6. Tastes*

| | |
|---|---|
| Favorite food | Cybernetic Pizza (Italian cuisine), Mechanical Sushi (Japanese cuisine) and Electronic Croissants (French cuisine). |
| Favorite drink | Robotics Orange Juice |
| Favorite color | Blue, like the sea and the sky. |
| Favourite animal | Dolphin (he is fascinated by it since it cannot swim). |
| Favourite book | The divine Comedy. |
| Favourite song | Bohemian Rhapsody. |
| Favorite device | Tablet (has one with it) |
| Favorite software | Android |
| Favorite Landscape | Sea. |
| Favourite sport | Running (since it cannot run it is fascinated by it) |

*Table 7. Goals - Short Term*

| |
|---|
| It wants to be of help to all those who need it. |
| It wants to make friends and have lots of new friends. |

*Table 8. Goals - Long Term*

| |
|---|
| It wants to learn new languages, because where it lived it always spoke Italian and English. |
| It wants to visit Florence because it is Dante's hometown. |
| It wants to build a hat with solar panels to self-recharge. |
| It wants to learn how to grasp objects. |
| It wants to learn how to play the piano. |

## 4. Conclusion and future work

As described in the analysis reported in [2], we realized that the robot personality and its character should be better defined and detailed in order to satisfy the curiosity and the expectations of children, and to make it more credible and engaging. At the moment, the robot often does not know how to answer the questions or does not understand them, and this generates friction and nervousness in children. So, we realized that we need to work more on its dialogue strategies and on enriching its knowledge. At the dialogue level, giving the robot a richer personality that it can convey through language, would constrain the expectations of the children to a more tractable set of topics, and create justifications for its inadequacies. At the current stage, the software architecture of the robot in our project encompasses a specialized module for casual conversation with autistic children, developed based on a set of experimental interactions with the children in real settings, which supports the believability of the robot character as an expert in a set of restricted domains through the use of external knowledge sources. The current architecture does not include a reinforcement learning component, but this element may be integrated in the robot to gain advantage from the monitoring of the responses of the children during the use of the robot on the field.

A possible direction for future work concerns the personalization of the robot generated messages. Indeed, previous work showed the possibility to guide the user toward a virtuous behaviour by using some specific word category (e.g.

adjectives, adverbs) and some specific sentence coordination strategy [1]. As future work, we are also working on the adaptive mechanisms with which Pepper will be enriched. The robot will show more and more intelligence and reasoning skills, which will allow it to adapt to the user's needs and customize the interaction in both adaptive and adaptable form. Thus, the robot will become an adaptive robot with respect to the user, able to adapt its behavior based on the characteristics of the subject with whom it is interacting and to customize the interaction in an adaptive form, that is, configurable by the staff expert in therapy. However, from our initial investigation we believe that the interaction should be "robot-guided" in order to give smarter answers to the (big variety) of questions collected.

## 5. Acknowledgements

The multidisciplinary project is carried out by the University of Turin, the Paideia Foundation, Intesa Sanpaolo Innovation Center and Jumple, and was funded by Banca Intesa Sanpaolo, Banca dei Territori Division.